# A batching and scheduling optimisation for a cutting work-center: Acta-Mobilier case study

(presented at the 7th IESM Conference, October 11 – 13, 2017, Saarbrücken, Germany)

Emmanuel ZIMMERMANN[1, 2, 3], Hind BRIL EL HAOUZI[1, 2], Philippe THOMAS[1, 2]
André THOMAS[1, 2], Mélanie NOYEL[3]

[1]Université de Lorraine, CRAN, UMR 7039, Campus Sciences, BP 70239, 54506 Vandœuvre-lès-Nancy cedex, France
[2]CNRS, CRAN, UMR7039, France
(e-mail: {emmanuel.zimmermann ; hind.el-haouzi ; philippe.thomas ; andre.thomas }@univ-lorraine.fr)
[3]Acta-Mobilier, parc d'activité Macherin Auxerre Nord 89270 MONETEAU
(e-mail : mnoyel@acta-mobilier.fr)

*Abstract*—The purpose of this study is to investigate an approach to group lots in batches and to schedule these batches on Acta-Mobilier cutting work-center while taking into account numerous constraints and objectives. The specific batching method was proposed to handle the Acta-Mobilier problem and a mathematical formalisation and genetic algorithm were proposed to deal with the scheduling problem. The proposed algorithm has been embedded in software to optimise production costs and emphasis the visual management on the production line. The application is currently being used in Acta-Mobilier plant and shows significant results

## I. Introduction

Acta-Mobilier, a high-quality product lacquering manufacturer, is faced to the high standards of quality and high technological requirements challenges which lead this company to have a reworks rate upper than 30% and reaching 80% for some products. Many critical consequences, including production cost increasing, products flow perturbations and steady deterioration of deliveries rate, are implied by this fact. Penalty costs and brand image damage may also be at stake.

In addition, a mass customisation business strategy has been adopted by the company, which leads to a significant diversified products panel and, at the same time, pushed to optimise its manufacturing processes.

To reduce the production cost and time and also answer a specific customer need a particular clustering/batching method on two levels combined to a scheduling optimization has been applied. In section 2, the industrial context is set. The section 3 exposes the studied problem. Then, the proposal is described in section 4. The obtained results are presented in the last section. The way these works would be used in a more global project is also described.

## II. Industrial context

The Acta-Mobilier company has two core business: front manufacturing for kitchen specialists (subcontracting) and the design and realization of shops and stands (layout). Both use the same human and material resources.

In this paper, we focused on the subcontracting business. To manage this activity the company has adapted the mass customization approach which leads to very customized products. Thus, each front has custom dimensions, thickness, colours and many other parameters like, for example, the handle dimensions and position. Consequently, the company have to deal with a big variability on customer demands and variability on manufacturing processes: different rooting sheets, rework rate… while keeping manufacturing process as standard and an optimised production cost.

In order to raise a high quality of product and in particular by avoiding differences in the texture and the appearance of the fronts which have to form a coherent and harmonious finished product reference, the company tried to manage its production by finished product reference (a finished product reference represents all the fronts forming a kitchen). For example, each employee in charge of applying the cement finishing has his own way to do it: the sense of application, thickness, painter claw .... There is also a risk of tint difference if all the fronts are not processed on the same period and with the same preparation, the drying time can also impact on the tint.

Moreover, sending the demand orders already sorted by finished product reference leads to a significant time saving for customer. All these requirements are essential regarding the company quality and customer service levels. For all those reasons, it becomes important to manufacture all fronts belonging to the same finished product reference on the same time.

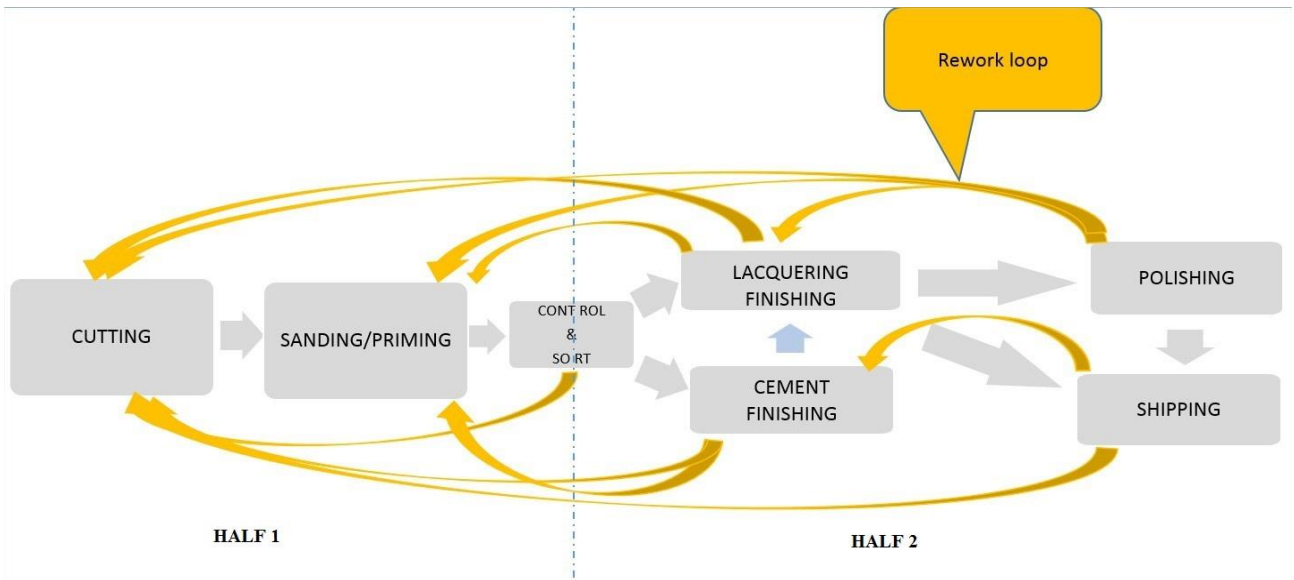

Fig. 1 Acta-Mobilier production line diagram

However, this policy requires to increase the surface area and implies many issues on material and time management (tool changes and organization in the cutting work-center). Consequently, the company needs a way to minimize time and production costs inferring the least possible functioning complication.

In previous works [15], the presence of indivisible batches, named "Kernels" has been highlighted. A Kernel is a batch of products of the same customer, with the same colour and having the same process (all the Kernel parts must stay together during the whole production line). These batches characteristics were particularly well adapted to the finishing process (the half 2 of Fig.1). The way to determine the Kernels composition is not the purpose of this work and should be seen as already known.

Working with this subdivision is irrelevant in the cutting work-center considering its process and objectives. That's why another batches construction regulation must be used to determine cutting batches: The cutting machines have dimensional limits lower than the ones available to customers and the fronts that exceed these limits are treated separately on another machine.

Moreover, the presence of a control step placed before the second part (Half 2 of Fig.1) of the manufacturing process will allow to split the cutting batches into the Kernels so that they could be worked one by one. To do so, it is, at a certain time, needed to physically achieve an unbundling and a re-bundling of the Kernels. However, the available floor space needed to achieve this sorting doesn't allow to have more than 6 or 7 pallets simultaneously. Regarding that it is impossible to increase this space, the batching organization has to take this into account.

## III. Problem statement and mathematical models

From the industrial context description, the cutting work-center optimizer problem can be seen as the combination of two well-known distinct problems classes:

- Making manufacturing batches in order to optimise the production costs without making too complex batches (increase space area, adding no-added value tasks to sort those batches in the end of the production line).
- Propose a schedule of proposed batches to optimise the control step. In this step, the human operator will be able to re-build the finished product references in less than an hour on the available floor space.

### A. Kernels batching

This problem aim is to find relevant manufacturing batches to minimise production and time wastes. There are at least two ways to interpret this problem:

- Take the whole customers' orders and split them into optimal sized lots which corresponds to the lot-sizing problems class [1], [2].
- Cluster the Kernels into bigger batches with common characteristics corresponding to the part family problems class [3], [4], [5].

In this study, the problem concerns the batches composition (Which distinct finished product references must be gathered? and how many batches?) and not the optimal number of products to put in batches. Moreover, as hypothesis, having the Kernel size already calculated, lead to put the studied problem in family part problems class. This class looks for grouping products that have the same relevant characteristics (e.g. thickness and production range).

Numerous works about solving this kind of problems exist, most of them use clustering techniques: [6] treated this kind of problem in order to minimize sequence dependent changeover times. In the company case, the changeover times are sequence independent but there is a strong constraint on the number of finished product reference to put in a batch. [3], [4], [5] worked on clustering with multi-criteria. Those works show that batching problems can totally differ from one production to another. In our case, although, the number of Kernel in each batch is not a strong constraint, the standard deviation must stay weak to not create disproportionate batches and smooth the production load. Many method could be used to solve batching problem for example [7] used a genetic algorithm to determine batches.

### B. Scheduling

The second problem is a batch scheduling problem on a single machine with multi criteria. The objective is to minimise the number of setup, but also to keep as close as possible the batches that are from the same finished product reference. The first objective is clearly to reduce production time and the second one guarantees that, on the control step, the collaborator will not wait more than one hour to have all the products he needs to rebuild the corresponding finished product references with enough floor space to achieve it well.

Many works were done on the subject of scheduling problems with setup times, [8] expose them. According to the classification of scheduling problem with setup times proposed by [9] presented in fig.2, the studied problem belongs to batch sequence independent scheduling problem.

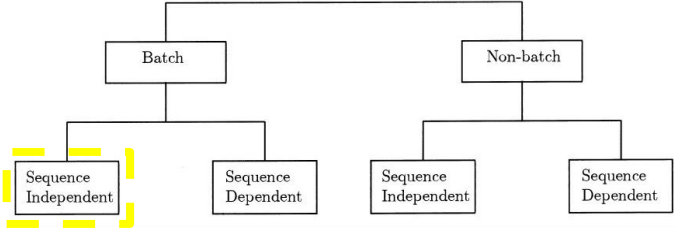

Fig. 2 The sscheduling problem with setup time problem classification [9]

To be more specific it belongs to the particular case of batch scheduling on a single machine, several solving methods were developed like branch and bound algorithm [10], Tabu search and also genetic algorithms [11],[12],[13].

A mathematical formulation of this specific problem is proposed below:

$N$ : $Number\ of\ finished\ product\ references$
$M$: $Number\ of\ finished\ product\ references\ batches\ (FPRB)$
$F$ : $Number\ of\ manufacturing\ batches\ (MB)$
$P_m$ : $Number\ of\ MB\ in\ FPRB\ number\ m$

The first function F1 has as objective to optimise the distance between all the $MB$ called $l_{e,m}$ of the same $FPRB$ and characterised by its thickness $e$ and its $FPRB$ number $m$:

$$F1 = \min_i \left( \sum_{j=1}^{F} a_{i,j}\ \max(x_j - x_i - 1, 0) \right)$$

Where:
- $a_{i,j}$ expresses the fact that the two $MB$ $i = l_{e,m}$ $and$ $j = l'_{e',m'}$ belong to the same $FPRB$

  $a_{i,j} = 1$ if m = m'
  $\quad\quad\quad 0\ elsewhere.$
- $x_i$ $is\ the\ position\ of\ MB\ i\ in\ the\ schedule$

The second function F2 has as objective to optimise the number of setup for the considered schedule:

$$F2 = \min_i \sum_{i=1}^{F} \sum_{j=1}^{F} y_{i,j}$$

Where:
- $y_{i,j} = 1\ if\ (e \neq e'\ \&\ j\ is\ right\ after\ i$
  $\quad\quad\quad 0\ elsewhere.$

Consequently, the global function to minimise is:
$F = \alpha\ F1 + \beta F2$, with α, β mixing $ratios$
This optimisation has to be done under the following constraints:

$F1 \geq 0$
$F2 \geq M$
$\forall i, x_i \in (0, F)$
$\forall (i,j),\quad y_{i,j} \in [\![0,1]\!]$
$\forall i, \quad \sum a_{i,j} = P_k\ where\ k\ is\ FPRB\ number\ of\ MB\ i$
$\forall j, \quad \sum a_{i,j} = P_k where\ k\ is\ FPRB\ number\ of\ MB\ j$

### IV. Proposal

### A. Clustering

To answer the objective of quality requirements and to provides an understandable and achieve a visual management standards for the human operators, the clustering has been made on two levels:
- On the first level, the different finished product references that have the more distinct thicknesses in common are gathered while also respecting the limit of grouping five finished references at most. This finished product references (FPR) batches are represented by an alphabetical character (e.g. batch A)

- On the second level the Kernels of each FPR batches with the same thickness are grouped into manufacturing batches, represented by the FRP batch character plus a randomly chosen integer (e.g. A1). The Fig.3 illustrated this explanation.

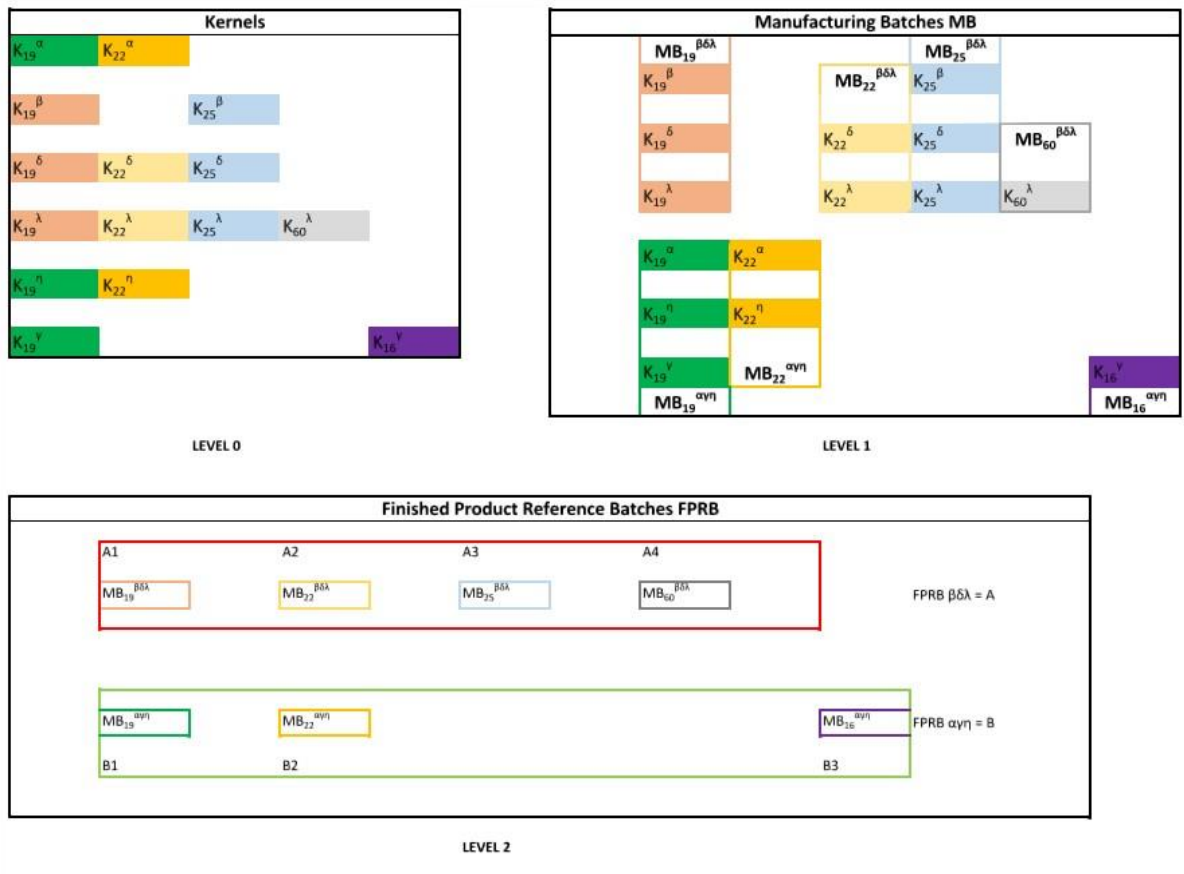

Fig. 3 : The two levels batching concept

With this two-levels batching, it is assured that the control & sort step will have enough floor space to reconstruct all the FPR Kernels without letting the pallets opened for a long time and without a lot no-added value tasks.

B. Scheduling

The chosen solving approach is the genetic algorithm because of their capacity to provide good results in competitive computing times [8]. In the studied case, the computation time is more important than having an optimal solution, because the company suffers from its high reworks rate susceptible to happen anywhere in the production line. Furthermore, the probability to be able to follow exactly the schedule is weak and a wider objective will be to re-calculate the schedule each time a disturbance appears.

Here it is proposed to apply a GA presented on Fig. 4 with the following parameters:

- An initial population of 1000 sequences of the batches randomly sorted.

- The natural selection with an elitism rule was chosen because even if, it is not the better rule to get closer to an optimal solution [14], it is known to be the fastest to converge and like already said velocity is more important than optimality.

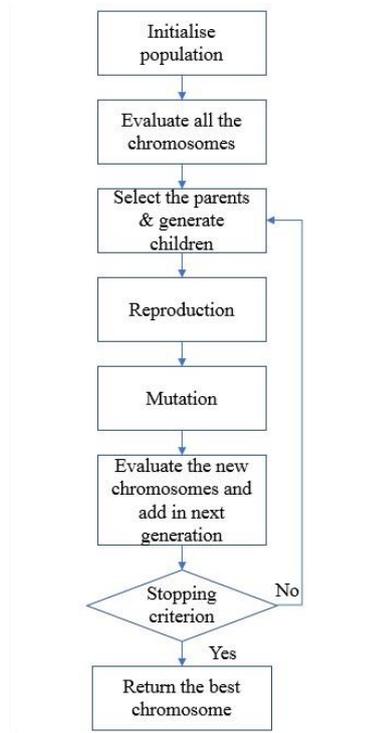

Fig. 4 A genetic algorithm flowchart

The crossover process is similar to [12] and is functioning by iteration:
Following the rule presented on Fig 5 a new population is created by mixing the chromosomes of the remaining sequences of the last generation:
In the first loop, the two sequences that best suits to the optimisation criterium are chosen to make the crossover. Then, in the second loop, two of the three best sequences are randomly picked for the crossover. And, it keeps incrementing until the new population reaches one thousand inhabitants.

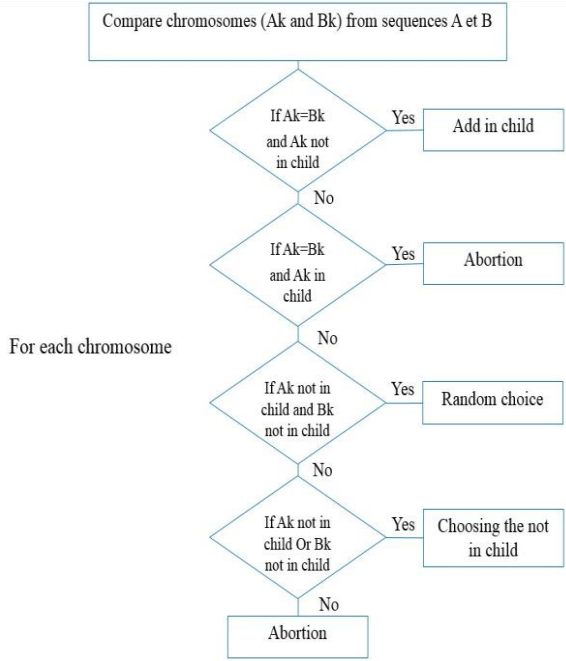

Fig. 5 Chromosome selection in crossover process

The mutation consists on swapping two chromosomes from a sequence. Two swapping methods were chosen: neighbour and foreign swapping explained on Fig. 6.
In a scheduling, it is non-sense to find the same batch at two distinct positions. That's why, in both mutation and crossover processes, the "abortion" phenomenon is important because it assumes that the same chromosome is not displayed more than a single time in a sequence.

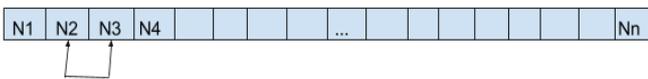
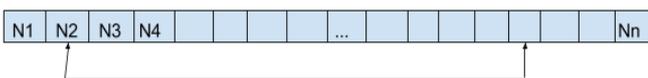

Fig. 6 Mutation principle

## V. Results

The proposed method was implemented with VB.NET and used to manage one customer demands since last year. As results, its application allowed the company to improve the relationship and increase the volume of demands from this customer: it was able to satisfy his quality, delivery date requirements and also reduce total production time by decreasing the addition of a non-value-added operation. Figure 7, presents the comparison of the performances of the cutting work-center before and after the deployment of the proposed solution on the same period.

In comparison to the previous year at the same period, with a 32% ordered quantity bigger and similar conditions, the global production time is one day faster. The results are presented in Table 1.

TABLE 1: A comparison with and without optimisation

|  | 2015 (without optimisation) | 2016 (with optimisation) |
|---|---|---|
| Number of fronts ordered by week | W49: 456 pcs<br>W50: 247 pcs<br>W51 596 pcs | W49: 633<br>W50: 518<br>W51: 561 |
| Number of days between the opening of the first and the last pallet in shipping workstation | W49: 5 days<br>W50: 5 days<br>W51: 5 days | W49: 4 days<br>W50: 4 days<br>W51: 4 days |

Those results show tangible benefits to the company but not lead to direct link between results and the application of the proposed method. In order to validate the efficiency of the proposed method, we simulate one production week of the cutting work-center with a demand of 36 finished product references and 4 different thicknesses.

The results presented on Table 2 show the comparison between our method and two scheduling rules previously used by the company:
- Grouping all kernels from the same finished product reference into the same batch,
- Grouping all kernels with the same thickness into the same batch.

TABLE 2: A comparison between the proposed approach and the old used scheduling rules

|  | Proposed method | Finished product references | Thickness |
|---|---|---|---|
| Number of setups | 11 | 4 | 22 |
| Maximal $W_{ip}$ between Kernels of the same FPR | 5 | 14 | 4 |
| Maximal number of pallets opened simultaneously | 7 | 16 | 2 |

In addition, the developed application leads to a solution easily understandable by the co-workers thanks to its visual representation. Also, specifics documents are elaborated for the cutting WorkCentre and others for the rest of the process.

On the control step, a specific visual application has been developed to help the co-worker to simply split the manufacturing batches (named with the letters) and regroup together the Kernels by their FPR, the whole functioning using the coloured code shown in Fig.7.

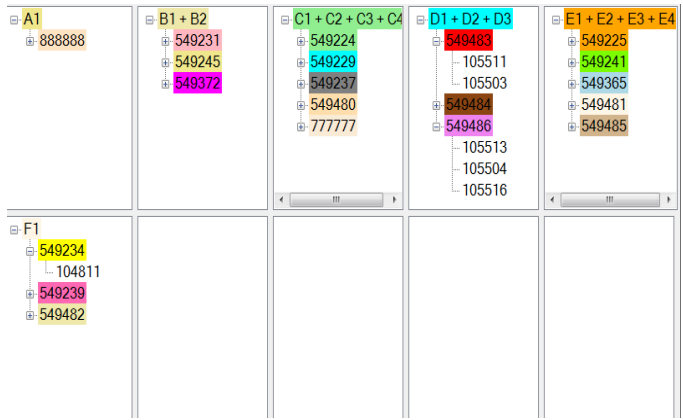

Fig. 7 An application screenshot presenting how to re-sort batches

## VI. Outlooks

In this paper, an efficient method to batch and schedule the manufacturing orders for a specific problem induced by Acta-mobilier company was presented. Even if of the significant results presented in this study, on-going work aims to use a CRAN laboratory platform named Tracilogis to determine the robustness of the generated solutions in case of reworks. Furthermore, this work can be seen as a part of wider project aimed to propose a hybrid manufacturing system which has been presented in [15]. The objective will be to use the developed solution as a local optimiser and make it collaborate with the other optimisers to provide the company a global dynamic system able to react to any disruption.